# Indoor PM$_{2.5}$ forecasting and the association with outdoor air pollution: a modelling study based on sensor data in Australia

Wenhua Yu[1], Bahareh Nakisa[2], Seng W. Loke[2], Svetlana Stevanovic[3], Yuming Guo[1], Mohammad Naim Rastgoo[2],

**Affiliations:**

[1]Climate, Air Quality Research Unit, School of Public Health and Preventive Medicine, Monash University, Level 2, 553 St Kilda Road, Melbourne, VIC 3004, Australia.

[2]School of IT, Faculty of Science and Built Environment, Deakin University, Australia

[3]School of Engineering, Deakin University, Waurn Ponds 3216, Australia

**\*Corresponding authors:**

Bahareh Nakisa, bahar.nakisa@deakin.edu.au

School of IT, Faculty of Science and Built Environment, Deakin University, Australia

**Declaration of competing financial interests:** The authors declare that they have no actual or potential competing financial interests.

**Abstract**

Exposure to poor indoor air quality poses significant health risks, necessitating thorough assessment to mitigate associated dangers. This study aims to predict hourly indoor fine particulate matter (PM2.5) concentrations and investigate their correlation with outdoor PM2.5 levels across 24 distinct buildings in Australia. Indoor air quality data were gathered from 91 monitoring sensors in eight Australian cities spanning 2019 to 2022. Employing an innovative three-stage deep ensemble machine learning framework (DEML), comprising three base models (Support Vector Machine, Random Forest, and eXtreme Gradient Boosting) and two meta-models (Random Forest and Generalized Linear Model), hourly indoor PM2.5 concentrations were predicted. The model's accuracy was evaluated using a rolling windows approach, comparing its performance against three benchmark algorithms (SVM, RF, and XGBoost). Additionally, a correlation analysis assessed the relationship between indoor and outdoor PM2.5 concentrations. Results indicate that the DEML model consistently outperformed benchmark models, achieving an R2 ranging from 0.63 to 0.99 and RMSE from 0.01 to 0.663 µg/m3 for most sensors. Notably, outdoor PM2.5 concentrations significantly impacted indoor air quality, particularly evident during events like bushfires. This study underscores the importance of accurate indoor air quality prediction, crucial for developing location-specific early warning systems and informing effective interventions. By promoting protective behaviors, these efforts contribute to enhanced public health outcomes.

**Introduction**

Household air pollution poses substantial health risks to the public. According to the World Health Organization, it was implicated in approximately 3.2 million premature deaths in 2020 (World Health Organization 2022). Fine particulate matter, specifically those with a diameter of 2.5 micrometres or smaller ($PM_{2.5}$) represents one of the most harmful air pollutants to residential health. Existing evidence indicates that indoor $PM_{2.5}$ is positively associated with asthma, acute respiratory infection, lung cancer, and tuberculosis (Lee et al. 2020). Most individuals, especially those in high-income countries, are estimated to spend approximately 90% of their time in indoor environments (Klepeis et al. 2001). The indoor air quality thus emerges as a pivotal factor influencing personal health. Prolonged exposure to poor indoor air quality intensifies the impact of air pollution on individual health. Therefore, evaluating indoor air quality is essential to minimize exposure to high levels of air pollution exposure and assess the associated health risks.

Indoor air quality can be influenced by various factors, including ventilations, energy sources, human behaviours, construction styles and materials, and cultural practices, resulting in complex relationship with ambient air pollution (Mannan and Al-Ghamdi 2021). The concentrations of $PM_{2.5}$ in indoor and outdoor environments are interconnected. Outdoor $PM_{2.5}$ can infiltrate indoor spaces, especially for buildings where its locations closed to outdoor air pollution sources, such as traffic-busy roads, industrial facilities, and bushfire events. Buildings located near these air pollution sources may experience higher infiltration rates of outdoor $PM_{2.5}$, resulting in elevated indoor concentrations. Additionally, indoor levels of $PM_{2.5}$ can sometimes exceed outdoor levels, particularly when indoor air pollution sources are present. A range of indoor sources, including wood and coal fires, as well as cooking stoves, can contribute to higher indoor $PM_{2.5}$ concentrations than the outdoor level (Alastair C. Lewis 2023). Therefore, understanding the association between indoor and outdoor $PM_{2.5}$ is significant to identify the source of pollutants and further inform mitigation strategies and preventative measures.

The widespread use of low-cost smart air quality monitors provides an opportunity to monitor the indoor air quality through building automation systems (Haase et al. 2016). The long-term air quality data sets, in conjunction with advanced machine learning (ML) algorithms, offer the potential to predict future indoor air quality. Accurate prediction of indoor air quality can result in more energy-efficient operation of heating, ventilation, and air conditioning (HVAC) systems, and identify the sources and causes of indoor air pollution, preventing timely the onset of health issues related to poor air quality. In this study, we aim to identify the association between indoor and outdoor $PM_{2.5}$ concentrations and forecast hourly indoor $PM_{2.5}$ concentrations using data from 91 sensors across 24 commercial and residential buildings in Australia. We also introduce the development of a sophisticated air quality monitoring dashboard. Designed for real-time data analysis in multi-level buildings, the dashboard integrates sensor networks to track key air quality parameters. Its notable feature is the ability to generate detailed reports on demand, offering insights into the environmental conditions of a selected building over a specified period.

**Methods**

**Data collection**

We collected the indoor environmental data from 91 indoor air quality monitoring sensors across 21 commercial and three residential buildings in Australia. The indoor hourly temperature, and relative humidity, $PM_{2.5}$, Particulate Matter 10 micrometres or less in diameter ($PM_{10}$), total volatile organic compounds (TVOC) were measured through the low-cost, cloud-based calibrated air monitoring sensors from the August 2019 and November 2022. The details about the installation and calibration of the sensors can be found in our previous study (Yu et al. 2023). Briefly, the monitoring sensors were strategically mounted on interior walls at a height of 1.5 to 1.7 meters from the ground. Optimal sensor placement was determined based on supply and return air on each floor, deliberately avoiding potential indoor air pollution sources and vents such as windows, mechanical HVAC systems, and kitchens. Prior to installation, professional metrology laboratories conducted inter-comparison tests on all sensors, and annual calibrations and uncertainty estimations were performed and documented. All commercial buildings had their air conditioning systems regularly operating during business hours. In contrast, the air in residential buildings was not regulated by air conditioning systems and were subject to natural fluctuations. The details about the number of sensors in each of 24 building locations and recordings periods are presented in Table 1. It should be noted that we included one building in Symonston, ACT in Australia, where it suffered a direct influence on the 'Black Summer' bushfire events from December 2019 to February 2020 (Yu et al. 2020).

We obtained the outdoor meteorological data from the EAR5 dataset (the fifth-generation Medium-Range Weather Forecasts European Centre for Reanalysis set) (Hersbach et al. 2020), including hourly outdoor temperature, ambient dew point temperature (2 meters above the land surface), wind speeds at a height of 10 meters above sea level, surface pressure, solar radiation, and total precipitation in 2019 - 2022. The outdoor climate data were then linked with the geographic locations of each sensor.

The ambient monitoring station-based hourly $PM_{2.5}$ data were collected from the regional Environmental Protection Agencies (EPAs) in Australia (Centre for Air pollution energy and health Research 2021; Riley et al. 2020). Since the station based hourly $PM_{2.5}$ data were only available until December 2020, we limited the comparison analyses between outdoor and indoor hourly $PM_{2.5}$ within the period from 2019 to 2020. Ultimately, there were nine commercial buildings involved in the indoor and outdoor $PM_{2.5}$ comparison analyses.

**Data analysis**

We used a deep ensemble machine learning (DEML) framework to estimate the hourly indoor $PM_{2.5}$ concentration. The details of the DEML framework have been discussed in our previous study (Yu et al. 2022; Yu et al. 2023). In brief, three base models [Random Forest (RF), eXtreme Gradient Boosting (XGBoost), and Support Vector Machine (SVM)] were used at the first stage to estimate the indoor hourly $PM_{2.5}$ for each sensor. In the second stage, two meta-models [RF and Generalized Linear Model (GLM)] were trained using the combination of the predictions generated in the first-stage base models. Finally, a nonnegative least squares (NNLS) algorithm was used in the third stage to obtain the optimal weights of meta-model results and to achieve the ultimate indoor $PM_{2.5}$ forecasting. We used three benchmark models including RF, SVM, and XGboost to compare the performance with the proposed DEML. The specific details of these benchmark models can be found elsewhere (Alpaydin 2020; Bishop 2006). All model establishments were developed using R software (version 3.5.3).

We established DEML model for each sensor to predict the hourly mean indoor $PM_{2.5}$ concentrations using the collected indoor environmental indictors, including hourly indoor $PM_{10}$, TVOC, temperature, and relative humidity, as well as outdoor climate and air pollution measurements, such as ambient temperature and dew point temperature, wind speeds, surface pressure, solar radiation, total precipitation, and outdoor $PM_{2.5}$ concentrations at the corresponding sensor locations. Considering the potential for a robust autocorrelation between future hourly indoor $PM_{2.5}$ and those observed in preceding hours, we also incorporated the indoor $PM_{2.5}$ from previous 3-6 hours and from the corresponding values on the prior day. Additionally, the model integrated the geographical coordinates of the structures housing the sensors, along with temporal data, encompassing the year, month, day, day of the week, hour, and season. We assigned the last 10% of days for each sensor as the testing dataset and the remaining days as the training dataset.

For each time-series sensor data, a moving window technique was employed to assess the stability of the DEML model over time and prevent overfitting (Yu et al. 2023). To be specific, a fixed 24-hour "moving window" (referred to as the forecast horizon) was established within the testing dataset. The forecast horizon is then moved forward one day per step to fit the DEML model and predict the indoor $PM_{2.5}$ levels within the forecast horizon. After the prediction of the indoor $PM_{2.5}$ in the first forecast horizon, this forecast horizon data was added to the training set in the next step, and another DEML model will be refitted using the enlarged training data to predict indoor $PM_{2.5}$ in the following forecast horizon. The process continued until predictions were generated for the entire testing data. The amalgamation of predictions within the testing dataset was computed to evaluate the model performance. We calculated the root mean square error (RMSE) and the coefficients of determination ($R^2$) to assess the model performance.

The Spearman correlation analysis was conducted to investigate the correlation between indoor and outdoor $PM_{2.5}$. The smooth plots were used to illustrate the consistency of observed indoor and outdoor $PM_{2.5}$ in study city. To examine the variables' contribution in indoor $PM_{2.5}$ forecasting, both RF and XGBoost models were utilized to determine the variables' importance. The loss of RMSE was calculated to measure how much a model's performance changes if the effect of a selected explanatory variable is removed. 10 permutations were selected to repeat the process 10 times to compute the mean values of RMSE loss.

Results

Table 1 presents the descriptive statistics of hourly average $PM_{2.5}$ concentrations ($\mu g/m^3$) collected from 24 buildings in Australia between 2019 and 2022. The hourly $PM_{2.5}$ concentrations displayed considerable variations across different locations, with the minimum concentrations of 0.00 $\mu g/m^3$ to the maximum concentrations of 87.06 $\mu g/m^3$. The median $PM_{2.5}$ concentrations in the study periods were 1.11 $\mu g/m^3$ and three locations in NSW (Macquarie and Sydney) and QLD (Brisbane) experienced a slightly higher indoor $PM_{2.5}$ above 2.06 $\mu g/m^3$. The interquartile range (IQR) of $PM_{2.5}$ concentrations varied between locations, with the lowest value of 0.00 $\mu g/m^3$ observed in Greenway, ACT, and the highest value of 4.56 $\mu g/m^3$ in Curtin, ACT.

Table 2 summarizes the indoor hourly $PM_{2.5}$ forecasting performance of the DEML model and three benchmark models (RF, XGBoost, and SVM) in the study buildings in Australia. Overall, the DEML model demonstrated superior performance in forecasting indoor hourly $PM_{2.5}$ concentrations compared to the benchmark models. In most cases, the DEML model achieved higher $R^2$ values and lower RMSE values than the other models, with $R^2$ of 0.63 - 0.99 and RMSE of 0.01 - 0.663 $\mu g/m^3$, indicating better prediction accuracy and lower prediction errors.

**Table 1.** The descriptive statistics for the hourly average PM$_{2.5}$ (µg/m$^3$) in 91 sensors in 24 buildings from 2019 to 2022 in Australia

| state | suburb | Start time | End time | Sensors | Study days | Study hours | Min | Max | Median | IQR | Mean | SD |
|---|---|---|---|---|---|---|---|---|---|---|---|---|
| ACT | Conder | 26/03/2022 | 21/11/2022 | 2 | 241 | 11,468 | 0.00 | 76.37 | 0.02 | 2.23 | 1.91 | 4.04 |
| ACT | Curtin | 19/04/2022 | 21/11/2022 | 2 | 216 | 9,621 | 0.95 | 81.57 | 2.48 | 4.56 | 4.85 | 7.60 |
| ACT | Deakin | 16/03/2020 | 21/11/2022 | 5 | 980 | 95,341 | 0.00 | 73.67 | 0.95 | 1.26 | 2.31 | 3.16 |
| ACT | Fyshwick | 9/06/2020 | 21/11/2022 | 4 | 895 | 42,741 | 0.92 | 80.99 | 0.95 | 0.88 | 1.91 | 2.56 |
| ACT | Greenway | 18/09/2022 | 21/11/2022 | 32 | 64 | 49,875 | 0.00 | 17.50 | 0.00 | 0.00 | 0.29 | 1.02 |
| ACT | Parkes | 17/03/2020 | 21/11/2022 | 5 | 979 | 134,508 | 0.39 | 62.59 | 0.95 | 0.11 | 1.60 | 2.18 |
| ACT | Symonston | 3/08/2019 | 2/01/2022 | 6 | 882 | 37,396 | 0.00 | 87.06 | 0.95 | 0.38 | 2.81 | 8.15 |
| ACT | Symonston | 28/07/2020 | 21/11/2022 | 4 | 846 | 68,216 | 0.00 | 39.10 | 0.95 | 0.68 | 1.64 | 2.00 |
| NSW | Macquarie | 1/03/2021 | 21/11/2022 | 3 | 630 | 29,226 | 0.95 | 84.86 | 2.17 | 4.42 | 4.47 | 6.07 |
| NSW | Macquarie | 13/12/2021 | 21/11/2022 | 2 | 343 | 10,598 | 0.95 | 58.40 | 0.95 | 0.27 | 1.44 | 1.59 |
| NSW | Newcastle | 9/02/2022 | 21/11/2022 | 1 | 285 | 6,396 | 0.00 | 17.74 | 0.95 | 0.14 | 1.37 | 1.21 |
| NSW | Parramatta | 27/01/2020 | 21/11/2022 | 2 | 1029 | 20,802 | 0.95 | 51.72 | 0.96 | 0.99 | 1.93 | 2.49 |
| NSW | Sydney | 14/12/2021 | 21/11/2022 | 1 | 342 | 7,279 | 0.00 | 59.94 | 2.06 | 3.62 | 3.58 | 3.96 |
| NSW | Sydney | 1/02/2020 | 21/11/2022 | 4 | 1025 | 25,626 | 0.83 | 82.77 | 0.98 | 0.99 | 2.14 | 3.42 |
| QLD | Brisbane | 13/02/2020 | 21/11/2022 | 3 | 1012 | 26,417 | 0.00 | 38.02 | 0.95 | 0.00 | 1.19 | 1.27 |
| QLD | Brisbane | 1/07/2022 | 21/11/2022 | 1 | 143 | 3,085 | 0.95 | 33.20 | 1.71 | 2.56 | 2.95 | 3.17 |
| QLD | Brisbane | 6/05/2022 | 21/11/2022 | 1 | 199 | 2,377 | 0.95 | 29.32 | 1.17 | 1.35 | 2.06 | 1.98 |
| SA | Adelaide | 3/03/2020 | 21/11/2022 | 1 | 993 | 11,655 | 0.00 | 64.63 | 1.00 | 0.84 | 1.71 | 1.89 |
| SA | Adelaide | 15/03/2022 | 21/11/2022 | 2 | 251 | 6,397 | 0.86 | 27.25 | 0.95 | 0.07 | 1.22 | 0.84 |
| VIC | Bendigo | 18/05/2022 | 21/11/2022 | 1 | 187 | 3,018 | 0.87 | 30.39 | 0.95 | 0.66 | 1.82 | 2.06 |
| VIC | Melbourne | 5/09/2019 | 1/11/2022 | 4 | 1152 | 63,911 | 0.05 | 84.72 | 1.11 | 1.81 | 2.63 | 3.72 |
| VIC | Melbourne | 2/02/2022 | 21/11/2022 | 1 | 292 | 6,597 | 0.95 | 26.06 | 1.18 | 1.21 | 2.25 | 2.62 |
| VIC | Melbourne | 3/02/2022 | 21/11/2022 | 2 | 291 | 10,619 | 0.00 | 26.14 | 0.95 | 0.40 | 1.94 | 2.67 |
| VIC | Rowville | 23/05/2022 | 21/11/2022 | 2 | 182 | 6,183 | 0.00 | 59.05 | 1.46 | 3.90 | 3.22 | 4.83 |

Note: PM$_{2.5}$: particles with a diameter of < 2.5 µm; SD: Standard deviation; IQR: the interquartile range of PM$_{2.5}$ concentrations in the study period.

**Table 2.** Indoor hourly PM$_{2.5}$ forecasting performance of DEML model and three benchmark models from 24 buildings in Australia.

|  |  |  | RF | | XGBoost | | SVM | | DEML[a] | |
| --- | --- | --- | --- | --- | --- | --- | --- | --- | --- | --- |
| State | Suburb | Test hours | R2 | RMSE | R2 | RMSE | R2 | RMSE | R2 | RMSE |
| ACT | Symonston* | 843 | 0.956 | 0.392 | 0.976 | 0.309 | 0.840 | 0.745 | **0.974** | **0.299** |
| ACT | Parkes* | 535 | 0.995 | 0.073 | 0.996 | 0.070 | 0.989 | 0.128 | **0.997** | **0.057** |
| ACT | Deakin* | 420 | 0.987 | 0.051 | 0.997 | 0.033 | 0.941 | 0.155 | **0.998** | **0.020** |
| ACT | Fyshwick | 465 | 0.959 | 0.435 | 0.983 | 0.403 | **0.986** | **0.350** | 0.977 | 0.394 |
| ACT | Symonston* | 336 | 0.999 | 0.007 | 0.999 | 0.013 | 0.931 | 0.124 | **0.999** | **0.010** |
| ACT | Greenway* | 2768 | 0.503 | 0.045 | 0.621 | 0.033 | 0.140 | 0.091 | **0.631** | **0.030** |
| ACT | Conder* | 318 | 0.508 | 0.113 | 0.938 | 0.039 | 0.164 | 0.289 | **0.931** | **0.037** |
| ACT | Curtin* | 313 | 0.969 | 0.817 | 0.996 | 0.384 | 0.984 | 0.636 | 0.994 | **0.355** |
| NSW | Sydney | 391 | 0.888 | 0.286 | **0.992** | **0.078** | 0.918 | 0.227 | 0.987 | 0.091 |
| NSW | Parramatta | 154 | 0.914 | 0.049 | **0.943** | **0.036** | 0.315 | 0.160 | 0.927 | 0.044 |
| NSW | Macquarie* | 252 | 0.991 | 0.380 | 0.995 | 0.268 | 0.987 | 0.382 | **0.995** | **0.244** |
| NSW | Sydney | 82 | 0.985 | 0.164 | **0.993** | **0.109** | 0.982 | 0.221 | 0.992 | 0.127 |
| NSW | Macquarie* | 241 | 0.977 | 0.073 | 0.984 | 0.067 | 0.969 | 0.087 | **0.985** | **0.058** |
| NSW | Newcastle* | 83 | 0.951 | 0.025 | 0.986 | 0.016 | 0.420 | 0.094 | **0.987** | **0.013** |
| QLD | Brisbane | 322 | **0.959** | **0.164** | 0.907 | 0.255 | 0.949 | 0.197 | 0.924 | 0.232 |
| QLD | Brisbane* | 83 | 0.921 | 1.552 | 0.988 | 0.692 | 0.989 | 0.859 | **0.990** | **0.626** |
| QLD | Brisbane* | 82 | 0.980 | 0.307 | 0.993 | 0.253 | 0.988 | 0.324 | **0.994** | **0.190** |
| SA | Adelaide* | 86 | 0.983 | 0.021 | 0.994 | 0.018 | 0.704 | 0.093 | **0.983** | **0.021** |
| SA | Adelaide | 166 | 0.608 | 0.052 | **0.997** | **0.014** | 0.330 | 0.073 | 0.910 | 0.021 |
| VIC | Melbourne | 586 | 0.958 | 1.003 | **0.999** | **0.378** | 0.974 | 0.870 | 0.995 | 0.413 |
| VIC | Melbourne* | 84 | 0.985 | 0.059 | 0.987 | 0.069 | 0.739 | 0.246 | **0.990** | **0.049** |
| VIC | Melbourne* | 244 | 0.990 | 0.232 | 0.997 | 0.154 | 0.993 | 0.192 | **0.997** | **0.111** |
| VIC | Bendigo | 83 | 0.249 | 0.018 | **0.999** | **0.013** | 0.004 | 0.151 | 0.415 | 0.024 |
| VIC | Rowville | 290 | 0.952 | 0.227 | **0.982** | **0.133** | 0.930 | 0.286 | 0.978 | 0.145 |

Note: PM$_{2.5}$: particles with a diameter of < 2.5 μm; R$^2$ is the coefficients of determination for unseen independent data; RMSE: the root-mean-square error; SVM: support vector machine; RF: Random Forest; XGBoost: extreme gradient boosting; DEML: the three-stage stacked deep ensemble machine learning method. *a* was a three-level structure where four base models (GBM, SVM, RF, and XGBoost) and three second-level models (RF, XGBoost, and GLM) were constructed with an NNLS algorithm at the third level. "*" and bold font indicate the DEML achieved the highest model performance compared with the other five benchmarks.

Table 3 displays the distribution and correlation between hourly mean indoor and outdoor $PM_{2.5}$ in nine buildings across Australia from 2019 to 2020. We observed that the mean concentrations of outdoor hourly $PM_{2.5}$ were more than two times higher than the corresponding indoor concentrations, ranging from $4.80 \pm 2.56$ µg/m³ to $9.91 \pm 19.61$ µg/m³. The highest mean indoor and outdoor $PM_{2.5}$ values were recorded in Symonston, ACT ($4.83 \pm 12.65$ µg/m³ VS. $9.91 \pm 19.61$ µg/m³), which experienced a bushfire event in 2019-2020. The correlation between indoor and outdoor $PM_{2.5}$ levels varied across the different locations. The strongest correlations were observed in Parramatta, NSW ($r = 0.70$), Symonston, ACT ($r = 0.69$), and Melbourne, VIC ($r = 0.69$), while the weakest correlation was identified in Brisbane, QLD ($r = 0.23$). Our analysis revealed that high outdoor $PM_{2.5}$ concentrations (hourly mean $PM_{2.5}$ above 6 µg/m³) tended to correspond with stronger correlation coefficients ($r \geq 0.69$), suggesting a more pronounced influence of outdoor $PM_{2.5}$ levels on indoor air quality under elevated outdoor pollution conditions. Figure 1 displays the temporal trends of sensor-based indoor $PM_{2.5}$ and station-based outdoor $PM_{2.5}$ in nine involved buildings in Australia. We observed that indoor $PM_{2.5}$ concentrations generally exhibited highly consistent temporal trends with outdoor $PM_{2.5}$, with the exception of the building in Brisbane, QLD (Figure 1).

In addition, we assessed the contributions of various factors within the RF and XGboost models. We found that the most important variables for hourly indoor $PM_{2.5}$ concentrations include $PM_{10}$, indoor $PM_{2.5}$ from the preceding one and two hours, and the building locations. The outdoor $PM_{2.5}$ concentrations had the fifth ranking in indoor $PM_{2.5}$ estimations in the XGboost Model. Details on the contributions of the top 15 variables can be observed in Figure 2.

**Table 3**. The correlation between hourly mean indoor $PM_{2.5}$ and station-based outdoor hourly $PM_{2.5}$ in 9 involved buildings from 2019 to 2020 in Australia.

| State | Suburb | Start time | End time | Indoor $PM_{2.5}$ | | Outdoor $PM_{2.5}$ | | $r^a$ |
|---|---|---|---|---|---|---|---|---|
| | | | | Mean | SD | Mean | SD | |
| ACT | Symonston[b] | 2019-08-03 | 2021-01-01 | 4.83 | 12.65 | 9.91 | 19.61 | 0.69 |
| ACT | Parkes | 2020-03-17 | 2021-01-01 | 1.76 | 1.64 | 5.10 | 2.85 | 0.42 |
| ACT | Deakin | 2020-03-16 | 2021-01-01 | 2.57 | 2.93 | 5.13 | 2.86 | 0.55 |
| ACT | Fyshwick | 2020-06-09 | 2020-11-20 | 2.17 | 2.31 | 5.17 | 2.80 | 0.46 |
| ACT | Symonston | 2020-07-28 | 2021-01-01 | 1.82 | 1.49 | 4.80 | 2.56 | 0.50 |
| NSW | Sydney | 2020-02-01 | 2021-01-01 | 1.63 | 2.03 | 5.57 | 5.37 | 0.52 |
| NSW | Parramatta | 2020-01-27 | 2021-01-01 | 2.70 | 3.56 | 6.26 | 5.67 | 0.70 |
| QLD | Brisbane | 2020-02-13 | 2020-03-23 | 0.91 | 0.24 | 5.48 | 1.70 | 0.23 |
| VIC | Melbourne | 2019-09-05 | 2021-01-01 | 2.57 | 5.54 | 6.63 | 6.28 | 0.69 |

Notes: $PM_{2.5}$: particles with a diameter of < 2.5 µm; SD: SD: Standard deviation.

*a* the coefficient of Spearman correlation analysis

*b* indoor $PM_{2.5}$ were monitored in this building in Symonston during bushfire event in 2019-2020.

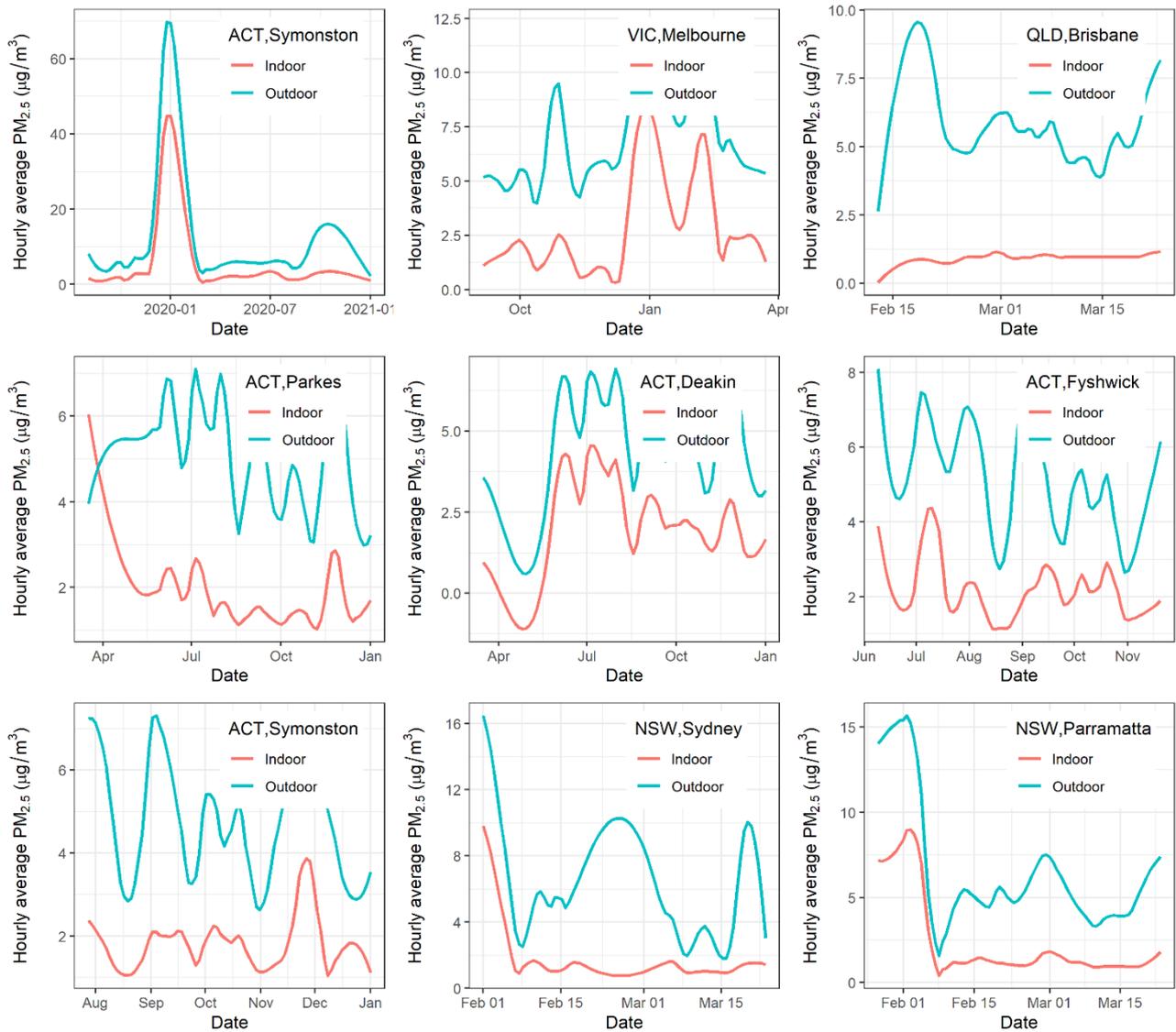

**Figure 1.** The comparison between sensor-based indoor PM$_{2.5}$ and station-based outdoor PM$_{2.5}$ in 9 involved buildings in Australia. The X-axis indicates the observed period for both valid monitor stations and indoor sensors; Y-axis indicates the hourly mean PM$_{2.5}$ observed by sensors or air quality monitors; The solid line represents a regression line for observed PM$_{2.5}$ with a smooth of 0.2. PM$_{2.5}$: particles with a diameter of < 2.5 µm.

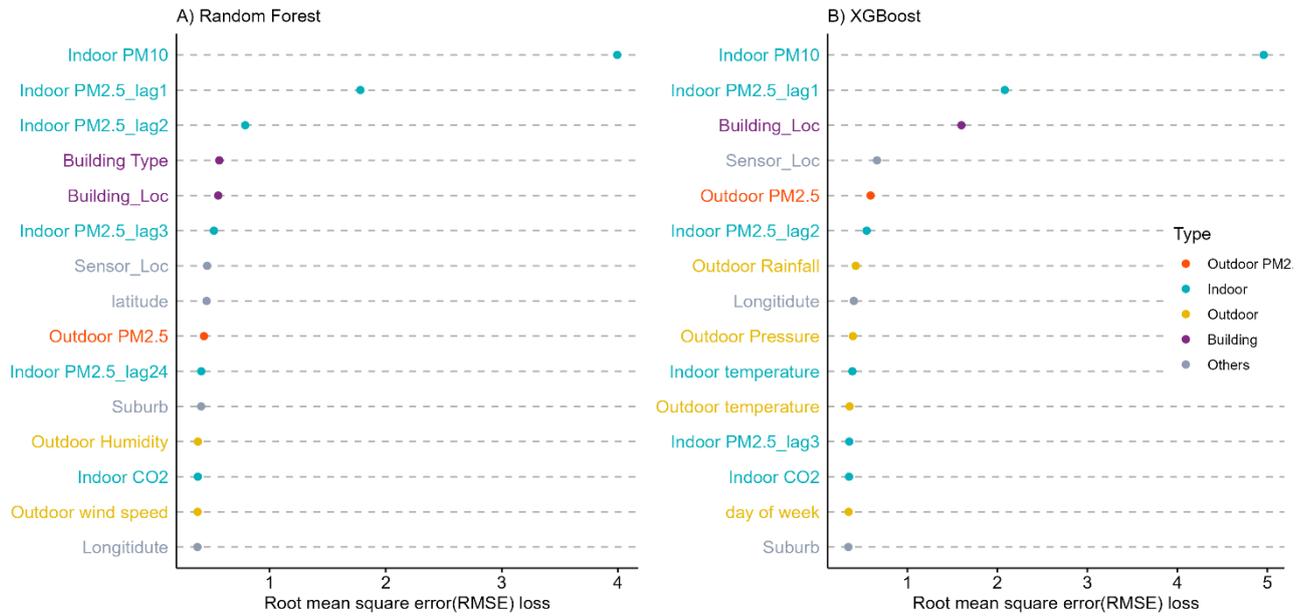

**Figure 2.** The contributions of 15 most important explanatory variables in indoor $PM_{2.5}$ estimation by RF and XGBoost in Australia.

Note: The loss of root mean square error was calculated to measure how much a model's performance change if the effect of a selected explanatory variable is removed. 10 permutations were selected to repeat the process 10 times to compute the mean values of RMSE loss. The RF and XGBoost models were used to calculate the variable importance. RF: Random Forest; XGBoost: extreme gradient boosting. $PM_{2.5}$: particles with a diameter of < 2.5 µm.

**A Real-time Air-Quality Monitoring Dashboard**

We developed an innovative dashboard tailored for the real-time monitoring and visualization of air quality indicators within a multi-tiered building structure. In selecting a platform for the dashboard development, Power BI was chosen due to its robust data processing capabilities and advanced visualization tools. Power BI's ability to seamlessly handle large datasets was crucial, given the volume and complexity of the air quality data collected from multiple sensors.

The developed dashboard integrates a graphical representation of each floor's plan, enhanced by the strategic deployment of air quality sensors across the building. These sensors are tasked with the measurement of a range of parameters, namely TVOC, PM10, PM2.5, CO2, as well as environmental factors like Temperature, Humidity, and Air Pressure. The dashboard presents these air quality metrics using a color-coded schema within a table format, which is based on established thresholds for each parameter. This design facilitates immediate recognition and interpretation of air quality data, enabling users to efficiently identify and address any deviations from standard air quality levels. This system thus provides an essential tool for maintaining and managing indoor air quality, ensuring compliance with health and safety standards.

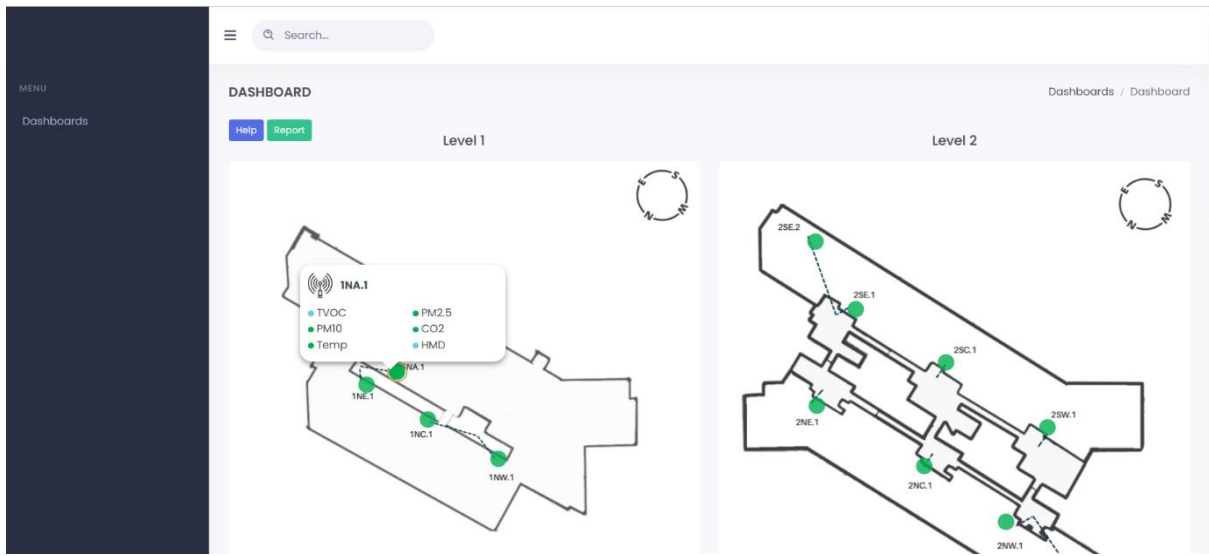

Figure 3. The developed air quality monitoring dashboard for a multi-tier building.

With this dashboard, we have improved the functionality of an air quality monitoring system by adding a feature for generating comprehensive reports. Users can access a list of sensor-equipped buildings and monitor their air quality through the dashboard. By choosing a particular building and setting specific start and end dates, the dashboard fetches and processes sensor data from the cloud, creating an in-depth report. This report offers critical insights into the air quality within the chosen building for the selected period.

**Discussion**

In this study, we deployed a cutting-edge DEML model to predict hourly indoor $PM_{2.5}$ concentrations, using data from 91 indoor air quality sensors distributed across 24 buildings throughout Australia. The DEML model demonstrated superior performance in predicting indoor $PM_{2.5}$ concentrations compared to the established benchmark models. The primary determinants for predicting indoor $PM_{2.5}$ included indoor $PM_{10}$, historical indoor $PM_{2.5}$ from preceding hours, geographical locations of the buildings, and outdoor $PM_{2.5}$ concentrations. Outdoor $PM_{2.5}$ concentrations were found to exert significant influence on the indoor air quality, especially in scenarios characterized by high outdoor air pollution.

A large body of studies has used machine learning and statistical models to predict indoor air quality (Bellinger et al. 2017; Wei et al. 2019). According a literature review, the most frequently used models to predict indoor $PM_{2.5}$ concentrations include artificial neural networks (ANN), various regression models, and decision trees-based algorithms (Wei et al. 2019). For example, Jorge Loy-Benitez and colleagues used a standard RNN, long short-term memory (LSTM), and a gated recurrent unit (GRU) structure of RNN separately to forecast hourly indoor $PM_{2.5}$ in subway stations in South Korea. They demonstrated superior performance with the GRU structure, achieving a RMSE of 21.05 ug/m$^3$ and $R^2 = 0.65$ (Loy-Benitez et al. 2019). Qi and colleagues used an exponential regression model to estimate indoor $PM_{2.5}$ concentrations in 13 households in Beijing, China. The authors used outdoor $PM_{2.5}$ with 80 minutes of lag time as the primary variable. The model reached a notable R2 of 0.861 in heating season (Qi et al. 2017). Yuchi et al used the random forest model and multiple linear regression for estimating indoor $PM_{2.5}$ concentrations from 342 apartments in Ulaanbaatar, Mongolia. The two model demonstrated comparable performance, with $R^2$ of 0.50 and 0.49 respectively (Yuchi et al. 2019). In comparison to previous studies, our DEML model achieved a notable high degree of accuracy in the prediction of hourly indoor $PM_{2.5}$ across most of the locations, with $R^2$ ranging from 0.63 to 0.99.

To improve the performance of models in indoor air quality prediction, one of the primary challenges encountered lies in the appropriate model selection. Given the diversity in ML algorithms, each model presents its own strengths and weaknesses, how to select the most suitable approach can be a complex task, and it often hinges heavily on empirical experience. A practical alternative strategy involves training a variety of models and subsequently integrating their predictive results, thereby deriving a more robust overall estimation (Wei et al. 2019). Hybrid modelling, such as the ensemble model, emerges as a promising technique, permitting the integration of various models to bolster the overall performance of the resultant model (Zhang et al. 2022). A wealth of studies in environmental exposure estimation have indicated that the ensemble approach by integrating multiple models can yield superior model performance compared to any individual ML used in isolation (Di et al. 2019; Yu et al. 2022). A particular study used a hybrid approach, blending a GLM model with RF predictions to forecast indoor $PM_{2.5}$ concentration. This innovative combination results in a better performance than the original GLM model alone, improving $R^2$ from 0.50 to a significant 0.82 (Yuchi et al. 2019). In our DEML model, we used a three-level stacked ensemble model (Yu et al. 2022), incorporating the capabilities of RF, XGBoost, SVM, and GLM algorithms to achieve optimal predictive indoor $PM_{2.5}$ performance. In line with previous studies, our findings indicated that the ensemble model surpass any constructed single model.

The model performance largely relies on the selection of predictors. According to a well-conducted literature review, indoor environmental indictors, including previous indoor $PM_{2.5}$, $PM_{10}$, indoor relative humidity, CO2 and NOx, as well as outdoor $pm_{2.5-10}$ and outdoor $PM_{2.5}$ are the most popular inputs for studies of indoor PM concentration predictions (Wei et al. 2019). Consistent with previous studies (Wei et al. 2019), we examined the most important variables in our study including indoor $PM_{10}$, previous indoor $PM_{2.5}$, and current outdoor $PM_{2.5}$.

The relationship between indoor and outdoor PM concentrations varies based on many factors including ventilation, construction type and age, infiltration rate of outdoor $PM_{2.5}$ into the indoor environment, weather patterns and meteorological conditions, and internal emission sources (Qi et al. 2017). Good ventilation system such as using central AC operation or window Fan operation is significantly associated with outdoor-to-indoor air exchange rate and the proportion of outdoor $PM_{2.5}$ in the indoor environment ($F_{INF}$) (Meng et al. 2009). In addition, outdoor temperatures and seasonal variation also related to the association between indoor and outdoor air pollution. A study found that the largest $F_{INF}$ was observed for the outdoor temperature with approximately 20 °C. Long et al found a seasonal pattern for indoor-outdoor air pollution fraction, with higher $F_{INF}$ in summer and lower in winter, when people tend to keep windows and doors closed to conserve heat (Long et al. 2001). Similarly, Qi et al also found significant differences in the correlation between indoor and outdoor $PM_{2.5}$ for the non-heating and heating season in China. In our study, we investigated the association between indoor and outdoor $PM_{2.5}$ in nine commercial buildings with regular AC operation. We found that the indoor $PM_{2.5}$ concentrations were generally lower than the outdoor concentrations, indicating the relatively reduced indoor $PM_{2.5}$ emissions. Moreover, we observed a variety of relationships between indoor and outdoor $PM_{2.5}$ across different locations, which experience distinct climate and weather conditions.

Indoor human activities such as smoking, heating, combustion, and cleaning have been reported as significant determinants of indoor $PM_{2.5}$ concentration (Meng et al. 2009). For example, smoking is considered one of the main sources of indoor air pollution, especially in high-income country (Ni et al. 2020). Indoor $PM_{2.5}$ concentration in dwellings with smoker were significantly higher than those in non-smoking buildings (Meng et al. 2009). Aside from actions that can pollute indoor air, there are preventive measures to reduce indoor $PM_{2.5}$ concentrations, such as opening windows and doors, utilizing fans and air conditioners, and implementing air freshener or filter. In this study, we did not take into account the individual activities variables on indoor air quality due to the lack of relevant data. Additionally, it is important to note that our study took place in Australia during the COVID-19 pandemic, a time when domestic human activities were substantially limited due to lockdowns and stay-at-home orders. As a result, people might have altered their usual home or working activities. For example, there could have been less time working in the office, more cooking as opposed to eating out or ordering takeout, more cleaning activities to reduce the potential spread of the virus, or no guests due to social distancing rules. All these changes could influence the indoor environment, including potential sources of indoor air pollution. However, further analysis is warranted to explore how COVID-19 restrictions have impacted indoor $PM_{2.5}$ levels.

Numerous studies have established that outdoor air pollution is the principal determinant and the largest contributor to indoor $PM_{2.5}$ levels, particularly in environments with limited internal emission sources. A study that analysed 374 non-smoking households in the USA found that outdoor $PM_{2.5}$ significantly contributed to indoor concentrations, accounting for 20% to 90% of the indoor levels. (Meng et al. 2009). Another study examined indoor and outdoor $PM_{2.5}$ concentrations simultaneously using on-line particulate counters across 13 households in Haidian, Beijing. Their findings revealed a high degree of correlations (> 0.9) between indoor and outdoor $PM_{2.5}$ (Qi et al. 2017). In our study, we also found that outdoor $PM_{2.5}$ was one of the most substantial factors in predicting indoor $PM_{2.5}$. In addition, environments with elevated levels of outdoor air pollution, such as during bushfire events, can lead to increased infiltration rates of outdoor $PM_{2.5}$, thereby affecting indoor air quality more significantly. Rajagopalan and Goodman found that bushfire smoke in Australia can substantially increase the levels of pollutants within residential buildings (Rajagopalan and Goodman 2021). Consistent with previous findings, our study also observed a consistent trend and higher correlations for buildings exposed to higher outdoor $PM_{2.5}$ concentrations. Among the buildings in this study, the one directly impacted by a bushfire event exhibited the greatest impact, with a correlation of 0.7.

Our study has many strengths. First, we included a large dataset which is sufficiently representative, encompassing a wide array of climate zones and indoor environmental conditions in numerous Australian urban areas. Our data come from 91 sensors in 24 buildings covering eight cities in Australia with a high temporal resolution (hourly) over a period of three years from 2019-2022. It considered not only the commercial and residential types of buildings, but also certain specific situations such as bushfire events. The DEML models achieved consistently high accuracy across different scenarios. Furthermore, we made several innovations in the indoor $PM_{2.5}$ concentrations prediction: 1) It is the first to implement an innovative DEML framework combining several ML approaches for future indoor $PM_{2.5}$ prediction. 2) We proposed a sliding windows approach in time-series data to improve temporal generalization and avoid over-fitting. 3) We used every sensor itself for indoor monitoring data and the corresponding outdoor environmental exposure data as inputs to establish the DEML model for each sensor, making our approach independent for different sensor locations. Finally, we examined the potential impact of ambient $PM_{2.5}$ on indoor air quality in various locations.

Several limitations in our study are warranted to discuss. First, even though we examined the building types, location, and the use of the AC operations, other building features such as building materials, surrounding greenery or air pollution resources may impact the variations of the indoor air pollution. Additionally, the efficiency of the ventilation system and human activities could also influence the indoor air variations and are warranted for further analysis. Furthermore, even though we installed several sensors even in the same building, spatial variability in indoor $PM_{2.5}$ levels may still exist from one room to another, which pose challenges for the representation of indoor $PM_{2.5}$ concentrations. Moreover, we collected the hourly outdoor $PM_{2.5}$ data from the nearest monitoring stations, while hourly monitoring station data after 2020 was unavailable. The actual outdoor $PM_{2.5}$ around the buildings may be different from the monitoring stations, which may lead to potential bias for the estimated associations between indoor and outdoor $PM_{2.5}$. In addition, the sensor-based indoor air quality data were mainly collected during the COVID-19 pandemic period in Australia, when the indoor and outdoor air quality may be influent by the modified human behaviours. Last, this study only focuses on the buildings in the metropolitan environments in Australia and the association between indoor and outdoor $PM_{2.5}$ may be different in rural or outskirt areas, where buildings are subject to distinct microclimate environments and ventilation conditions.

**Conclusion**

This study used a DEML model to predict the hourly indoor $PM_{2.5}$ concentrations across 24 buildings in eight Australian cities. Our results indicated that incorporating the capabilities of multiple levels of ML algorithms could achieve optimal predictive indoor $PM_{2.5}$ performance. We also observed that outdoor $PM_{2.5}$ concentrations significantly impact on indoor air quality, particularly for environments with elevated levels of outdoor air pollution, such as during bushfire events. These findings could help to develop location-specific indoor air quality early warning systems and inform more effective interventions or protective behaviours such as avoiding smoking indoors, using exhaust fans when cooking, or using air purifiers to reduce indoor $PM_{2.5}$ exposure and protect residential health. In addition, the variations across locations and times in pollution levels suggest the need for localized (fine-grained) sensor-based air pollution monitoring in real-time, indoors and outdoors.